\documentclass{article}
\usepackage{spconf,amsmath,graphicx,hyperref}
\usepackage{mathtools} 
\usepackage{booktabs}
\usepackage{makecell}
\usepackage{enumitem} 
\usepackage{siunitx}
\usepackage{booktabs,multirow,colortbl,xcolor}
\usepackage[T1]{fontenc}      % required for Type-1
\newcommand{\native}[1]{\cellcolor{orange!20}{#1}}   % light gray shade
\usepackage{tcolorbox}
\usepackage{float}        % put this in the preamble
\usepackage{amssymb}   % or: \usepackage{amsfonts}

\title{A Comparative Study on How Data Normalization Affects Zero-Shot Generalization in Time Series Foundation Models}

\name{Ihab Ahmed$^{\star\dagger}$ \qquad Denis Krompa{\ss}$^{\star}$ \qquad Cheng Feng$^{\star}$ \qquad Volker Tresp$^{\dagger}$}

\address{$^{\star}$ Siemens AG, Munich, Germany \\
         $^{\dagger}$ Ludwig Maximilian University of Munich, Germany \\
         \texttt{\{ahmed.ihab, denis.krompass, cheng.feng\}@siemens.com} \\
         \texttt{volker.tresp@lmu.de}}

\begin{document}
\maketitle
\ninept        % 9-point body text (required)

\begin{abstract}  
We investigate input normalization methods for Time Series Foundation Models (TSFMs). While normalization is well-studied in dataset-specific time series models, it remains overlooked in TSFMs where generalization is critical. Time series data, unlike text or images, exhibits significant scale variation across domains and channels, coupled with non-stationarity, can undermine TSFM performance regardless of architectural complexity. Through systematic evaluation using four architecturally diverse TSFMs, we empirically establish \textsc{RevIN} as the most efficient approach, reducing zero-shot MASE by 89\% relative to raw baseline and by 44\% in comparison to other normalization methods, while matching the best in-domain accuracy without any dataset-level preprocessing—yielding the highest accuracy-efficiency trade-off. Yet its effective utilization depends on architectural design choices and optimization objective, particularly with respect to training loss scale sensitivity and model type (point-forecast-, distribution- or LLM-based models).  
\end{abstract}
\section{Introduction}
\label{sec:intro}

Time Series Foundation Models (TSFMs) have emerged as an effective alternative to traditional one-model-per-dataset approaches in time series analysis. By pretraining models on large-scale, diverse time series datasets, TSFMs learn to capture complex temporal dynamics and enable knowledge transfer across domains. TSFMs have shown promising results in zero-shot forecasting and model generalization across various domains and unseen datasets, even challenging or exceeding the performance of full-shot models at times \cite{woo2024unified, cohen2025observability}.

TSFMs face unique challenge stemming from time series data’s heterogeneous, non-stationary nature \cite{wu2025outofdistributiongeneralizationtimeseries}. Scale and distribution shifts across domains and intra-domain features challenge generalization ability \cite{yao2025neuralscalinglawstime}. Hence, data normalization pivotally impacts TSFMs’ performance and generalization, yet systematic comparisons of normalization methods in TSFMs remain lacking despite advances in data normalization research in domain-specific settings \cite{kim2022revinn, han2024sin, NEURIPS2023_2e19dab9}. Current literature for TSFMs often adopts heuristic normalization choices without evidence-based justification.

Motivated by this gap, we conduct experiments to quantify data normalization's effect on TSFMs, and to answer questions: How do alternative normalization choices—specified by (i) the underlying statistics computed (mean – std, min – max, max-absolute, etc.) and (ii) the scope at which those statistics are measured (dataset-level versus instance-level)—influence the performance of TSFMs?, and what guidelines to follow for selecting and adapting normalization methods for optimal cross-domain adaptation?. Our contributions include: (1) a comparative analysis of data normalization in TSFMs, investigating its essential role in cross-domain generalization; (2) systematic evaluation of six normalization methods (\textsc{RevIN}, MeanAbs, Standardization+\textsc{RevIN}, Standardization, MinMax, MaxAbs) across four TSFMs (MOIRAI, CHRONOS, GTT, LightGTS) which span different architectural designs and training protocols, using multi-domain datasets spanning energy, transportation, and meteorology; (3) empirically showing that mean/std–based normalization methods consistently outperform others, improving TSFM forecasting accuracy relative to the raw baseline by approximately 89\% (zero‑shot) and 79\% (in‑domain), and by approximately 44\% (zero‑shot) and 28\% (in‑domain) relative to alternative normalization methods. (4) empirical insights into how \textsc{RevIN}'s effective utilization varies with model architecture design and optimization objective scale-sensitivity.

\section{Related Work}
\label{sec:related}

Zero-shot generalization in time series forecasting is challenged by non-stationarity and distribution shifts—a problem amplified in Time Series Foundation Models (TSFMs) by heterogeneous pretraining corpora that mix sampling rates and channel counts \cite{wu2025outofdistributiongeneralizationtimeseries, li2025nonstationarytimeseriesforecasting, liu2024moiraimoeempoweringtimeseries} as well as significant variation in numerical magnitudes across different datasets and channels \cite{salinas2020deepar, dooley2023forecastpfn, ansari2024chronos}. MOIRAI counters this with frequency-aware patch in/out projections, Any-variate Attention, and a mixture output head to handle arbitrary horizons and dimensions \cite{woo2024unified}. MOIRAI-MoE adds token-level sparse experts that specialize to regimes and frequencies without manual routing \cite{liu2024moiraimoeempoweringtimeseries}. Time-MoE scales sparse experts to billion-parameter capacity with multi-resolution heads, activating only a few experts per token \cite{shi2025timemoebillionscaletimeseries}. CHRONOS normalizes cross-dataset scale by performing instance scaling+quantization for each variate to a fixed vocabulary and then use a T5-style LLM backbone \cite{ansari2024chronos}. GTT predicts next curve shape using temporal attention as well as cross-channel attention \cite{feng2024curveshapematterstraining}. LightGTS aligns heterogeneous periods via cyclic tokenization and periodical decoding \cite{wang2025chattime}. Toto uses patch-wise causal scaling, factorized time–variate attention, and pretrain on a large mixed-frequency corpus \cite{cohen2025observability}. Lag-Llama conditions on explicit lag covariates for portable zero/few-shot use \cite{rasul2023lagllama}. Time-LLM reprograms frozen LLMs with prompt/prefix tuning  \cite{jin2024timellm}. MOMENT pretrains with masked patches on diverse corpora for frequency-agnostic transfer, and Chat-Time discretizes series for multimodal text-conditioned forecasting \cite{goswami2024momentfamilyopentimeseries, wang2025chattime}.

Input normalization is crucial for time series models, with studies showing its absence can degrade performance by 30\% \cite{passalis2019deepadaptiveinputnormalization} and amplify dominance of high-variance channels \cite{lima2023normalization, asesh2021normalization}. While traditional global methods like standardization \cite{zhou2021informer, liu2022scinet}, MinMax \cite{xi2024vsformer}, and MaxAbs \cite{maxabs} address input scale differences, they fail to resolve non-stationarity \cite{kim2022revinn}. Instance-level techniques like \textsc{RevIN} \cite{kim2022revinn}, SAN \cite{NEURIPS2023_2e19dab9}, FAN \cite{ye2024frequency}, and SIN \cite{han2024sin} mitigate distribution shifts via instance-specific statistics or adaptive parameterization. Despite advances, these methods are designed for and evaluated only in dataset-specific models, leaving open the question of their effectiveness in TSFMs under zero-shot demands. In addition, TSFM architects often adopt normalization heuristically, some provide little justification behind their choice, such as: Lag-Llama \cite{rasul2023lagllama} employs median/IQR scaling for outlier robustness, CHRONOS \cite{ansari2024chronos} uses instance mean scaling to preserve semantic zeros \cite{salinas2020deepar, rabanser2020effectivenessdiscretizationforecastingempirical}, and Chat-Time \cite{wang2025chattime} applies instance Min-Max for binning. However, other models default to \textsc{RevIN} without justification \cite{cohen2025observability, jin2024timellm, feng2024curveshapematterstraining, woo2024unified, das2024decoder, goswami2024momentfamilyopentimeseries, ekambaram2024tinytimemixersttms}.

Our work focuses on four representative TSFMs: MOIRAI (distribution forecasting), CHRONOS (LLM-based tokenization), and GTT/LightGTS (point forecasting)—the latter two offering distinct RevIN implementations critical to our normalization analysis (Section~\ref{ssec:pretrain}). We evaluate normalization methods: instance-level (\textsc{RevIN}, MeanAbs), dataset-level (Standardization, MinMax, MaxAbs), and a hybrid (Standardization→\textsc{RevIN}), with a raw baseline where no normalization is applied. While SAN/FAN/SIN fit single dataset models, their controls couple to data granularity: \textbf{FAN} normalizes via Top-K Fourier bins which depends on dataset's properties; Similarly, \textbf{SAN} utilizes hyperparameter T (slice length) which hinges on dataset's sample rate; \textbf{SIN} learns window-specific statistic sets and de-normalization maps that shift with the windowing/horizon distribution. Applying these methods to TSFMs is infeasible for two main reasons: (1) Inference: parameters like FAN's K or SAN's T cannot be determined for zero‑shot forecasting on previously unseen datasets/domains; (2) Training: estimating reliable, dataset‑dependent parameters during multi‑dataset pretraining is infeasible without novel mechanisms beyond our empirical scope. Hence, we do not consider SAN/FAN/SIN in this work.

%────────────────────────────────────────────────────────────
%  PRELIMINARIES – Normalisation Paradigms & Methods
%────────────────────────────────────────────────────────────
\section{Preliminaries}
\label{sec:prelim}

Let $\mathcal{D}= \{\mathbf{X}^{(1)},\dots,\mathbf{X}^{(n)}\}$ be
collection of multivariate time series
$\mathbf{X}^{(i)}\!\in\!\mathbb{R}^{T \times C}$ with $T$ time steps and
$C$ channels.  The observation at time step $t$ is
$\mathbf{x}^{(i)}_t\in\mathbb{R}^{C}$.
Normalization relies on two channel-wise vectors
$\boldsymbol{\beta}, \boldsymbol{\gamma}\in\mathbb{R}^{C}$ obtained from
shift and scale computed by functions $f_{\text{shift}}$ and $f_{\text{scale}}$. We consider two fundamental normalization paradigms:

\begin{itemize}[left=0pt]

  \item \textbf{Dataset-level:}  
         $\boldsymbol{\beta}$ and $\boldsymbol{\gamma}$ are calculated once from $\mathcal{D}$ and then applied to every observation $\mathbf{x}^{(i)}_t$:
        \begin{equation}
        \begin{aligned}
          \boldsymbol{\beta}^{\text{ds}}
          &= f_{\text{shift}}\!\bigl(\mathcal{D}\bigr), &
          \boldsymbol{\gamma}^{\text{ds}}
          &= f_{\text{scale}}\!\bigl(\mathcal{D}\bigr), \\[3pt]
          \tilde{\mathbf{x}}^{(i)}_t
          &= \frac{\mathbf{x}^{(i)}_t - \boldsymbol{\beta}^{\text{ds}}}
                  {\boldsymbol{\gamma}^{\text{ds}}}
        \end{aligned}\tag{1}
        \end{equation}

  \item \textbf{Instance-level:}  
         $\boldsymbol{\beta}$ and $\boldsymbol{\gamma}$ are computed for every look-back window
        $\mathbf{X}^{(i)}_{t-L:t}$ and then applied to every observation $\mathbf{x}^{(i)}_t$ in said window:
        \begin{equation}
        \begin{aligned}
          \boldsymbol{\beta}^{\text{inst}} &= f_{\text{shift}}\!\bigl(\mathbf{X}^{(i)}_{t-L:t}\bigr), 
          & \boldsymbol{\gamma}^{\text{inst}} &= f_{\text{scale}}\!\bigl(\mathbf{X}^{(i)}_{t-L:t}\bigr), \\
          \tilde{\mathbf{x}}^{(i)}_t &= \frac{\mathbf{x}^{(i)}_t - \boldsymbol{\beta}^{\text{inst}}}{\boldsymbol{\gamma}^{\text{inst}}}, 
        \end{aligned}
        \tag{2}
        \end{equation}

\end{itemize}

\noindent Given these definitions, normalization methods are defined as follows:

\begin{itemize}[left=0pt]
  \item \textbf{Dataset-level methods}
        \begin{itemize}[left=0pt]
          \item \emph{Standardization:}\;
                $\boldsymbol{\beta}^{\text{ds}}=
                \mu\bigl(\mathbf{X}^{(i)}_{1:T}\bigr),\;
                \boldsymbol{\gamma}^{\text{ds}}=
                \sigma\bigl(\mathbf{X}^{(i)}_{1:T}\bigr)$
          \item \emph{MinMax:}\;
                $\boldsymbol{\beta}^{\text{ds}}=
                \min\bigl(\mathbf{X}^{(i)}_{1:T}\bigr),\;
                \boldsymbol{\gamma}^{\text{ds}}=
                \max\bigl(\mathbf{X}^{(i)}_{1:T}\bigr)-
                \boldsymbol{\beta}^{\text{ds}}$
          \item \emph{MaxAbs:}\;
                $\boldsymbol{\beta}^{\text{ds}}=\mathbf{0},\;
                \boldsymbol{\gamma}^{\text{ds}}=
                \max\ \bigl|\mathbf{X}^{(i)}_{1:T}\bigr|$
        \end{itemize}

  \item \textbf{Instance-level methods}
        \begin{itemize}[left=0pt]
          \item \emph{RevIN:}\;
                $\boldsymbol{\beta}^{\text{inst}}=
                \mu\bigl(\mathbf{X}^{(i)}_{\,t-L:t}\bigr),\;
                \boldsymbol{\gamma}^{\text{inst}}=
                \sigma\bigl(\mathbf{X}^{(i)}_{\,t-L:t}\bigr)$
          \item \emph{MeanAbs:}\;
                $\boldsymbol{\beta}^{\text{inst}}=\mathbf{0},\;
                \boldsymbol{\gamma}^{\text{inst}}=
                \mu\ \bigl|\mathbf{X}^{(i)}_{\,t-L:t}\bigr|$
        \end{itemize}

  \item \textbf{Baseline (Raw):}\;
        no normalization
        ($\boldsymbol{\beta}=\mathbf{0},\;
         \boldsymbol{\gamma}=\mathbf{1}$).
\end{itemize}

\section{Methods}
\label{sec:print}

\begin{table}[H]          % capital H = absolutely here
\caption{Model Training Losses and Scale Sensitivity}
\label{tab:loss_scale}
\centering
\small
\begin{tabular}{l l l}
\toprule
\textbf{Model} & \textbf{Training Loss} & \textbf{Scale Sensitivity} \\
\midrule
MOIRAI          & NLL        & Invariant (distribution) \\
CHRONOS         & Token CE   & Invariant (pre-denorm) \\
GTT/LightGTS    & MAE/MSE    & Sensitive (raw scale)           \\
\bottomrule
\end{tabular}
\end{table}

\subsection{Time Series Foundation Models}
\label{ssec:tsfm}

Our experiment spans four TSFMs chosen for architectural and training loss diversity: MOIRAI~\cite{woo2024unified} (multivariate, distribution-based, Negative-Log-Likelihood loss), 
CHRONOS~\cite{ansari2024chronos} (univariate, LLM-based, token Cross-Entropy loss), 
GTT~\cite{feng2024curveshapematterstraining} (multivariate, point-forecast, MAE), and 
LightGTS~\cite{wang2025lightgts} (univariate, point-forecast, MSE). We use each model's smallest variant (Moirai-small, Chronos-tiny, GTT-tiny, LightGTS-tiny) to accommodate our pretraining corpus while isolating normalization effects from capacity differences. Crucially, each selected model's intrinsic design properties (Table~\ref{tab:loss_scale})—its output type (point vs. probabilistic forecasts) and training loss function scale-sensitivity during optimization—govern how normalization interacts with optimization dynamics: 

\begin{itemize}[left=0pt]
    \item \textbf{MOIRAI}: Outputs probabilistic distribution and optimizes NLL loss after de-normalizing the predicted distribution. Because the ground truth $(\mathbf{X} = \gamma^{\text{inst}} \,\tilde{\mathbf{X}})$ and predicted parameters $(\phi=\gamma^{\text{inst}} \,\tilde{\phi})$, have the same scaling factor $\gamma^{\text{inst}}$, the NLL decomposes as $\text{NLL}(\mathbf{X},\,\phi) = \text{NLL}(\tilde{\mathbf{X}},\,\tilde{\phi}) + \log(\gamma^{\text{inst}})$. Since $\log(\gamma^{\text{inst}})$ vanishes in the gradient ($\nabla_{\theta}\log(\gamma^{\text{inst}}) = 0$), the parameter update $\nabla_\theta\text{NLL}$ is invariant to input magnitude scaling, thus preventing optimization bias toward high-magnitude channels.

    \item \textbf{CHRONOS}: Computes cross-entropy loss on tokenized representations prior to instance de-normalization, where continuous values are discretized into scale-agnostic bins. By applying the loss to these normalized and quantized tokens (rather than de-normalized outputs), the optimization objective becomes inherently decoupled from input magnitudes—scale sensitivity is avoided by design through this tokenization-binning paradigm making the optimization objective scale-invariant.

    \item \textbf{GTT/LightGTS}: Output absolute values, optimizing scale-sensitive losses (MSE/MAE) on raw magnitudes. The MSE loss $\mathcal{L} = \frac{1}{C}\sum_{c=1}^C (x_c - \hat{x}_c)^2$ and MAE loss $\mathcal{L} = \frac{1}{C}\sum_{c=1}^C |x_c - \hat{x}_c|$ biases gradients $\nabla_\theta\mathcal{L}$ toward high-magnitude channels, necessitating strict shift/scale preservation across channels to prevent bias.

\end{itemize}

\subsection{Normalization in Pretraining}  
\label{ssec:pretrain}

Dataset-level normalization methods (Standardization, MinMax, MaxAbs) is straightforward to apply in all TSFMs, we preprocessed each dataset $\mathcal{D}$ with its own $\gamma^{\text{ds}}$ and $\beta^{\text{ds}}$ (Section~\ref{sec:prelim}) offline prior to training, and pretrain each TSFM without instance-level normalization, this is an effort to isolate the effect of dataset-level normalization vs instance-level normalization, especially when the statistics used are similar (Standardization vs \textsc{RevIN}).
However, instance-level methods utilization depends on design choices that are model-specific: Given MOIRAI's scale-invariant optimization, we retain its native \textsc{RevIN} implementation that de-normalizes the predicted distributions using $\gamma^{\text{inst}}$ and $\beta^{\text{inst}}$ before optimization step; CHRONOS optimizes cross-entropy loss on tokenized values before de-normalization; As established in section~\ref{ssec:tsfm}, the optimization objectives in GTT (MAE) and LightGTS (MSE) are
scale–sensitive when applied to raw values.  To mitigate this bias we employ the
instance–level normalization with clipping used in GTT for both
models.  Specifically, statistics $\gamma^{\text{inst}}$ and $\beta^{\text{inst}}$ are extracted from the context window
$\mathbf{X}_{t-L:t}$; these are then used to rescale the entire
instance—context and horizon $\mathbf{X}_{t-L:t+H}$—after which the loss is computed directly in the
normalized space:

\begin{align*}
    \gamma^{\text{inst}} &= f_{\text{scale}}\!\bigl(\mathbf{X}_{t-L:t}\bigr),
    &
    \beta^{\text{inst}} &= f_{\text{shift}}\!\bigl(\mathbf{X}_{t-L:t}\bigr),
    \\[2pt]
    \tilde{\mathbf{x}}_{\tau} &=
    \frac{\mathbf{x}_{\tau}-\beta^{\text{inst}}}{\gamma^{\text{inst}}},
    &
    \tau &= t-L,\dots,t+H.
    \tag{3}
\end{align*}

However, when the context scale is small ($\gamma^\text{inst}$ → $0$), normalized horizon values can become disproportionately large, so we discards outliers ($\left|\boldsymbol{\mathbf{X}}^{\text{inst}}\right| \le 10$). For GTT this is the original training protocol; LightGTS, by contrast, originally followed the native \textsc{RevIN} pipeline that normalizes only the context, de-normalizes the prediction
$\hat{\mathbf{X}}_{t+1:t+H}$, and optimizes MSE on raw scale (if no dataset-level standardization preceded).  We replace that procedure with the above clipped instance normalization, thereby making the objective scale-invariant, which significantly improves zero-shot forecasting accuracy as shown in table ~\ref{tab:revin_ablation}.

Lastly, many time series models (e.g., TimesNet~\cite{wu2023timesnet},
LightGTS~\cite{wang2025lightgts}) employ what we term \textsc{Hybrid}
normalization when referring to \textsc{RevIN}. We add this as a
separate method in our experiments to compare it with dataset-level standardization and instance-level \textsc{RevIN}: for every dataset $\mathcal{D}$ we first
apply dataset-level standardization using $\gamma^{\text{ds}}$ and
$\beta^{\text{ds}}$; during training we then normalize the context window $ \tilde{\mathbf{X}}_{t-L:t}$ using
$\gamma^{\text{inst}}$ and $\beta^{\text{inst}}$, following equations 1,2 from section~\ref{sec:prelim}, the process is defined as:
\begin{equation}
  \tilde{\mathbf{x}}_{t} =
  \frac{\mathbf{x}_{t} - \boldsymbol{\beta}^{\text{ds}}}
       {\boldsymbol{\gamma}^{\text{ds}}}
  \quad\text{then}\quad
  \tilde{\mathbf{x}}_{t} =
  \frac{\tilde{\mathbf{x}}_{t} - \boldsymbol{\beta}^{\text{inst}}}
       {\boldsymbol{\gamma}^{\text{inst}}}.
  \tag{4}
\end{equation}
the prediction is then de-normalized using the same $\gamma^{\text{inst}}$ and $\beta^{\text{inst}}$ prior to loss calculation and optimization (similar to native \textsc{RevIN}). Dataset-level normalization brings all channels to a comparable scale before training, so scale-sensitive objectives (e.g., MSE/MAE) becomes scale-invariant. Instance-level normalization then removes per-sample shift and scale to capture local non-stationarity; Consequently, even when the prediction is de-normalized with the instance statistics (as in native \textsc{RevIN}) the preceding dataset-level step ensures that gradients remain unbiased across channels.

% \noindent 1) \textbf{Dataset-level} puts all channels on a
% comparable scale before training, so a scale-sensitive optimization objective such as MSE or
% MAE becomes largely scale-invariant: no single high-magnitude channel can
% dominate the loss or its gradients. 

% \noindent 2) \textbf{Instance-level} then removes the
% sample-specific shift and scale inherent to each instance, capturing
% local non-stationarity; Consequently, even when the prediction is de-normalized with the instance statistics (as in native \textsc{RevIN}) the preceding dataset-level step
% ensures that gradients remain well balanced across channels.

% All training conducted on NVIDIA RTX A5000 GPUs for approx 100,000 steps per configuration.

\subsection{Evaluation Protocol}  
\label{ssec:eval}

When evaluating foundation models in a real-world setting, we assume that the input instance may come from an unseen dataset or domain, so normalization statistics can only be derived from the input context $\mathbf{X}_{t-L:t}$. For instance-level methods (\textsc{RevIN}, MeanAbs), this aligns naturally with training, however, for dataset-level methods, we replace $\gamma^{\text{ds}}$ and $\beta^{\text{ds}}$ with $\gamma^{\text{inst}}$ and $\beta^{\text{inst}}$ during inference (e.g., test-set MinMax uses $\min(\mathbf{X}_{t-L:t})$ and $\max(\mathbf{X}_{t-L:t}) - \min(\mathbf{X}_{t-L:t}$). In the Hybrid approach (Standardization$\rightarrow$\textsc{RevIN}), we retain only the \textsc{RevIN} component at inference time. We report Mean Absolute Scaled Error (MASE) as an evaluation metric to mitigate scale inconsistencies across normalization methods and bias toward high‑magnitude channels.

\begin{table}[t]
\caption{Datasets used and Evaluation Configurations.}
\label{tab:datasets}
\centering\footnotesize
\begin{tabular}{@{}l rr ll@{}}
\toprule
\textbf{Dataset} & \textbf{Train} & \textbf{Test} & \textbf{Freq.} & \textbf{Pred. Len.} \\ 
\midrule
Weather & 42,156 & 10,540 & 10min & 144 \\
Electricity & 134,995 & 5,261 & 15min & 96 \\
ETTh1/ETTh2 & 14,539 & 2,881 & 1h & 24 \\
ETTm1/ETTm2 & 58,159 & 11,521 & 15min & 96 \\
Traffic & 14,034 & 3,509 & 1h & 24 \\
Turkey Power & 17,544 & 8,760 & 1h & 24 \\
\bottomrule
\end{tabular}
\end{table}
\label{sec:illust}

\section{Experiments and Results}
% =============================================================
%  ZS + ID  table with cyan highlight for the model’s own
%  normaliser.  Bold = best, underline = 2nd best in each sub-row.
% =============================================================
\begin{table*}[t]
  \centering
  \label{tab:123}
  \caption{Aggregate MASE scores (lower = better).  
           ZS = zero-shot (mean ± std over three leave-one-dataset-out runs);  
           ID = in-domain (mean ± std over the same three variants, averaged
           over the test sets).}
  \label{tab:zs_id_combined_highlight}

  \setlength{\tabcolsep}{6pt}
  \begin{tabular}{
      l l
      *{6}{c}
      | c
  }
    \toprule
         & & \textsc{RevIN} & MeanAbs & Hybrid & Standard & MinMax & MaxAbs & Raw \\
    \cmidrule(lr){3-8}\cmidrule(lr){9-9}

    % ----------  Moirai  (native = RevIN)  ----------
    \multirow{2}{*}{\textbf{Moirai}}
      & ZS & \native{\textbf{1.09 ± 0.23}} & 1.30 ± 0.28 &
             \underline{1.14 ± 0.19} & 1.20 ± 0.27 &
             1.47 ± 0.56 & 1.68 ± 0.50 & 3.00 ± 1.39 \\
      & ID & \native{\textbf{0.85 ± 0.03}} & 0.94 ± 0.02 &
             \underline{0.87 ± 0.02} & 0.92 ± 0.04 &
             0.97 ± 0.04 & 1.04 ± 0.00 & 1.16 ± 0.11 \\
    \midrule

    % ----------  Chronos  (native = MeanAbs)  ----------
    \multirow{2}{*}{\textbf{Chronos}}
      & ZS & 0.85 ± 0.24 & \native{0.80 ± 0.18} &
             0.85 ± 0.26 & \textbf{0.78 ± 0.17} &
             \underline{0.79 ± 0.17} & 0.97 ± 0.30 & 11.31 ± 10.47 \\
      & ID & 0.72 ± 0.05 & \native{\textbf{0.68 ± 0.03}} &
             0.74 ± 0.04 & \underline{0.69 ± 0.05} &
             0.72 ± 0.03 & 0.74 ± 0.07 & 4.66 ± 1.74 \\
    \midrule

    % ----------  GTT  (native = RevIN)  ----------
    \multirow{2}{*}{\textbf{GTT}}
      & ZS & \native{\textbf{1.03 ± 0.47}} & 1.46 ± 1.01 &
             \underline{1.06 ± 0.49} & 1.07 ± 0.54 &
             1.38 ± 0.97 & 1.57 ± 0.98 & 3.41 ± 0.95 \\
      & ID & \native{0.73 ± 0.04} & 0.87 ± 0.04 &
             \textbf{0.71 ± 0.02} & \underline{0.72 ± 0.02} &
             0.85 ± 0.04 & 0.86 ± 0.07 & 1.52 ± 0.52 \\
    \midrule

    % ----------  LightGTS  (native = Hybrid)  ----------
    \multirow{2}{*}{\textbf{LightGTS}}
      & ZS & \textbf{1.11 ± 0.0} & 3.60 ± 2.29 &
             \native{\underline{1.13 ± 0.01}} & 1.19 ± 0.04 &
             1.24 ± 0.08 & 6.12 ± 5.00 & 19.82 ± 6.85 \\
      & ID & \underline{1.04 ± 0.03} & 2.58 ± 0.46 &
             \native{\textbf{1.03 ± 0.03}} & 1.06 ± 0.04 &
             1.22 ± 0.03 & 2.61 ± 0.78 & 8.76 ± 1.48 \\

    \midrule[1pt]

    % --------------------  Averages  ------------------
    \multirow{2}{*}{\textbf{Avg}}
      & ZS & \textbf{1.02} & 1.79 & \underline{1.05} & 1.06 & 1.22 & 2.58 & 9.38 \\
      & ID & \textbf{0.84} & 1.27 & \textbf{0.84} & \underline{0.85} & 0.94 & 1.31 & 4.02 \\
    \bottomrule
  \end{tabular}

  \vspace{0.6em}
  \footnotesize
  \textbf{Bold} = best, \underline{underlined} = second best in each sub-row;  Hybrid = Standard→\textsc{RevIN};
  Orange = normalization originally used by TSFM.
\end{table*}

\label{sec:results}

\subsection{Experimental Design}  
\label{ssec:setup}

Our benchmark consists of six cross-domain time series datasets (Table~\ref{tab:datasets}) commonly used in forecasting. 
We evaluate normalization methods under two typical foundation model scenarios: 
\textit{in-domain (ID)}, where the model is pretrained on data from the same domain as its test set, and 
\textit{zero-shot (ZS)}, where an entire dataset is excluded from pretraining. 
ZS thus becomes a more challenging setting, mirroring typical TSFM usage in which models must adapt to new domains with no prior domain exposure. All experiments adopt a 24-hour prediction horizon, based on each dataset’s frequency (e.g., 144 steps for Weather at 10-minute frequency). This ensures consistent prediction time spans across datasets while allowing us to focus on normalization comparisons rather than horizon lengths. For data balancing, we preserve each TSFM’s native approach wherever feasible: CHRONOS use per-dataset sampling probabilities that sum to 1 (e.g., 0.2 for Weather, 0.3 for Electricity), and in GTT we cap instances at 5{,}000 per dataset. While MOIRAI’s original codebase provides offline-calculated dataset proportions, we omit this mechanism in our implementation—similar to LightGTS—which does not include any dataset balancing logic.

Each TSFM is pretrained under seven normalization approaches, the six methods from Section~\ref{sec:prelim} plus a \emph{hybrid} method (Standardization→\textsc{RevIN}), using hyperparameters reported in the original work for each model.  For every method we train three leave-one-dataset-out variants, withholding Weather, Electricity, or Turkey Power for ZS testing, yielding $21$ pretrained models per TSFM.  Every run is optimized for $\sim\!100\text{k}$ steps on NVIDIA RTX A5000 GPUs; all datasets in Table~\ref{tab:datasets} participate in pretraining except the one withheld for the current variant.  At inference we evaluate each model (i) ZS on its withheld dataset and (ii) ID on all datasets seen during training, with the test-set for each dataset excluded from pretraining.  The same normalization scheme is reapplied, but with window statistics as prescribed in section~\ref{ssec:eval}. Our strategy is to compare normalization methods within architectures (each model acts as a controlled experiment). Comprehensive cross‑model performance can be seen in the original papers~\cite{woo2024unified, ansari2024chronos, feng2024curveshapematterstraining, wang2025lightgts}.
\subsection{Results and Discussion}  
\label{ssec:results_and_discussion}

The results of our experiments (Table~\ref{tab:zs_id_combined_highlight}) suggests that normalization methods using mean $\boldsymbol{\mu}$ and standard deviation $\boldsymbol{\sigma}$ as shift $\boldsymbol{\beta}$ and scale $\boldsymbol{\gamma}$—coupled with \textsc{RevIN} at inference time (Section~\ref{ssec:eval})—achieve superior performance in both zero-shot (ZS) and in-domain (ID) settings. In ZS evaluation, \textsc{RevIN} (1.02 MASE), Hybrid (1.05), and Standardization (1.06) outperform other methods (MinMax: 1.22, MeanAbs: 1.79, MaxAbs: 2.58), delivering an 89\% improvement over raw baseline (9.38 MASE) and 44\% improvement over normalization methods on average (1.86 MASE). For ID tasks, \textsc{RevIN}/Hybrid (0.84) and Standardization (0.85) show similar dominance, achieving 79\% improvement over raw baseline (4.02 MASE) and 28\% improvement over other normalization methods (MinMax: 0.94, MeanAbs: 1.27, MaxAbs: 1.31). CHRONOS forms a notable exception: while normalization significantly improved performance —producing 92.5\% and 84.6\% average gains over raw baselines in ZS and ID, respectively—the specific choice among the six normalization methods has little impact once data are scaled (±0.07 MASE variation in ZS and ±0.02 MASE in ID), we attribute this to its cross-entropy loss on discretized tokens (Section~\ref{ssec:tsfm}).

Even though in \textsc{RevIN} paper~\cite{kim2022revinn} two out of the three baselines they tested their method on—Informer~\cite{zhou2021informer} and SCINet~\cite{liu2022scinet}—  already use dataset-level
standardization and reported that adding \textsc{RevIN} (i.e., the hybrid method) improved performance significantly, those experiments were conducted on dataset-specific models. In our setting, a single TSFM is
pretrained on a multi-dataset corpus, and the benefit virtually vanishes:
the three mean/std-based methods—RevIN, Hybrid, and
Standardization—differ by at most ±0.03 MASE in ZS
and ±0.02 MASE in ID on average, this can suggest that, for data normalization in TSFMs, the choice of normalization statistics (mean–std vs.\ alternatives) has a greater impact on performance than the scope at which the statistics are computed. However we still see that \textsc{RevIN} is the most efficient choice out of the mean/std-based normalization methods: it matches their accuracy without a full‑corpus pass for dataset‑level standardization—and the duplicated, standardized copies it would potentially create—which is a significant saving especially for foundation models pretrained on massive amounts of data. \textsc{RevIN} instead computes window-level statistics online, giving it the best accuracy-efficiency trade-off. This advantage holds once instance normalization utilization is adapted to each architecture design and optimization objective (Sections~\ref{ssec:tsfm}~-~\ref{ssec:pretrain}): \textsc{RevIN} is the best ZS performer for MOIRAI (1.09 MASE), GTT (1.03), and LightGTS (1.11) establishing it as the most efficient go-to for ZS generalization.

\subsection{Ablation: Scale‑Sensitive Objectives Adaptation}
\label{ssec:scale_sensitive_adaptation}

To test whether adapting instance normalization is necessary for scale‑sensitive objectives, we re‑trained GTT and LightGTS (which optimize raw forecasts) with native RevIN and compared the results to the RevIN variant used in our reported results (Section~\ref{ssec:pretrain} and 
Table~\ref{tab:zs_id_combined_highlight}). The ablation shows that mitigating scale sensitivity substantially improves forecasting performance:

\begin{table}[H]
  \centering
  \caption{Effect of mitigating scale sensitivity: MASE}
  \label{tab:revin_ablation}
  \setlength{\tabcolsep}{10pt}
  \begin{tabular}{lccc}
    \toprule
    Model & Native & Clipped & Reduction \\
    \midrule
    LightGTS & 1.33 & \textbf{1.11} & \(16.5\%\) \\
    GTT      & 1.62 & \textbf{1.03} & \(36.4\%\) \\
    \bottomrule
  \end{tabular}
\end{table}

\section{Conclusion}
This work presents the first quantitative study of data normalization in Time Series Foundation Models (TSFMs). Across four architectures, dataset-, instance-, and hybrid normalization methods, and multi-domain benchmarks, we empirically find that mean/std-base normalization methods outperform others, reducing zero-shot MASE by 89\% over raw baseline and by 44\% over other normalization methods. \textsc{RevIN} delivers the best accuracy-efficiency trade-off: it matches dataset-level standardization yet avoids a full-corpus preprocessing pass—a major saving when pretraining large TSFMs on diverse corpus. Its benefit, however, depends on the optimization objective: scale-invariant losses accept \textsc{RevIN} directly, whereas scale-sensitive losses may require targeted adaptations. 
% By treating forecasting accuracy as a proxy for TSFM generalization, this analysis lays the groundwork for transferring these gains to classification, imputation, and anomaly detection.
% Because existing normalization research remains tied to dataset- or task-specific models, advancing temporal AI at scale demands TSFM-centric schemes that are dataset/task agnostic.
By using forecasting accuracy as a proxy for TSFM generalization, this work highlights the limitations of existing normalization research, which largely remains tied to dataset- or task-specific modeling approaches. Our findings indicate a clear need for dedicated, dataset- and task-agnostic normalization schemes tailored to TSFMs. While this study focuses on time series forecasting, the insights obtained may extend to other time series analysis tasks, such as classification, imputation, and anomaly detection, which we leave for future work.

\vfill\pagebreak
% BiBTeX files (here: refs). The IEEEbib.bst bibliography
\bibliographystyle{IEEEbib}
\bibliography{refs}

\end{document}